\documentclass{article}

\usepackage{PRIMEarxiv}

\usepackage[utf8]{inputenc} % allow utf-8 input
\usepackage[T1]{fontenc}    % use 8-bit T1 fonts
\usepackage{hyperref}       % hyperlinks
\usepackage{url}            % simple URL typesetting
\usepackage{booktabs}       % professional-quality tables
\usepackage{amsfonts}       % blackboard math symbols
\usepackage{nicefrac}       % compact symbols for 1/2, etc.
\usepackage{microtype}      % microtypography
\usepackage{lipsum}
\usepackage{fancyhdr}       % header
\usepackage{graphicx}       % graphics
\graphicspath{{media/}}     % organize your images and other figures under media/ folder

\usepackage{cite}
\usepackage{amsmath,amssymb,amsfonts}
\usepackage{algorithmic}
\usepackage{textcomp}
\usepackage{xcolor}

\title{The Generalizability of Explanations}

\author{
  Hanxiao Tan \\
  AI Group \\
  TU Dortmund \\
  \texttt{hanxiao.tan@tu-dortmund.de} }

\begin{document}
\maketitle

\newcommand{\beginsupplement}{%
        \setcounter{table}{0}
        \renewcommand{\thetable}{S\arabic{table}}%
        \setcounter{figure}{0}
        \renewcommand{\thefigure}{S\arabic{figure}}%
        \setcounter{equation}{0}
        \renewcommand{\theequation}{S\arabic{equation}}%
     }
     
\begin{abstract}
Due to the absence of ground truth, objective evaluation of explainability methods is an essential research direction. So far, the vast majority of evaluations can be summarized into three categories, namely human evaluation, sensitivity testing, and salinity check. This work proposes a novel evaluation methodology from the perspective of generalizability. We employ an Autoencoder to learn the distributions of the generated explanations and observe their learnability as well as the plausibility of the learned distributional features. We first briefly demonstrate the evaluation idea of the proposed approach at LIME, and then quantitatively evaluate multiple popular explainability methods. We also find that smoothing the explanations with SmoothGrad can significantly enhance the generalizability of explanations.
\end{abstract}

\section{Introduction}
As the performance of machine learning models has grown dramatically in recent years, they are increasingly being deployed in a wide range of fields. However, the potential dangers regarding trustworthiness are stepping into the limelight. For instance, in the medical domain, where decisions directly affect human lives, most black-box models are incapable of shouldering this responsibility. The prevailing solutions are twofold: inherently interpretable models and post-hoc explainability methods. For the former, interpretable models are typically simple in structure such that they are inadequate for decision-making tasks of growing complexity \cite{linardatos2020explainable}. Although there are researchers who argue that well-screened features facilitate both interpretability and performance \cite{rudin2019stop}, traditional machine learning appears to be incompetent in the image or high-dimensional visual tasks. Post-hoc explainability approaches do not have excessive requirements for the internal architecture of the model and can provide users with convincing explanations while assuring the decision performance.

However, the current issue with post-hoc explainability methods is that the quantitative evaluation is challenging. Early studies evaluated explanations by collecting a large number of human scoring (user studies) \cite{narayanan2018humans,alqaraawi2020evaluating}. While user studies are human friendly, these approaches are costly and the experimental results are unreproducible due to the subjective nature of humans \cite{petsiuk2018rise}. A viable alternative is quantitative evaluation, i.e. automating the computations to achieve objective assessments. However, the greatest obstacle to objective evaluations is the absence of ground truth explanations. Existing studies have proposed available evaluation methods which verify the properties that a reasonable explanation should possess \cite{bach2015pixel,adebayo2018sanity,hooker2019benchmark}. These properties can be divided into two categories: the sensitivity to 1) the perturbations of the inputs \cite{bach2015pixel} and 2) the randomization of the model parameters (salinity checks) \cite{adebayo2018sanity}. The former is the most popular candidate, and the assumption is that perturbing important features results in a rapid decline in prediction confidence, and vice versa. Nevertheless, skeptics argue that the forced perturbation corrupts the data distribution (which the model never learns) and therefore cannot be assessed with the prediction confidence \cite{hooker2019benchmark}. In addition, it has been considered that the flip-flop perturbation does not take into account the correlations between features \cite{shi2020modified}. The logic of the salinity test is that \emph{a plausible explanation is necessarily model dependent}. However, it is not sufficient as an evaluation metric for explanations: most of the model-related outputs can pass the salinity check, although they apparently do not serve as explanations.

This work proposes a novel approach to evaluate explanations from the perspective of generalizability. Our assumption is that a plausible explanation should share a proximate distribution with the original data and we train a generative model that is adequate for the complexity of the data and observe the performance of reconstructing the explanations. Our approach is intuitive, applicable to all saliency map-like explanations, and without requirements for the type of explainability methods (both for gradient- and perturbation-based). Our contributions are as follows:

\begin{itemize}
    \item We propose a novel method for evaluating explanations that validates whether they possess information consistent with the original data from the perspective of generalizability.
    
    \item We evaluate popular saliency map-based explainability methods.

    \item We show the interesting observation that SmoothGrad can effectively enhance the generalizability of explanations.
\end{itemize}

The structure of this paper is as follows: In Section \ref{RelatedWork} we present the relevant studies, in Section \ref{Methods} we detail the proposed approach, and in Section \ref{Experiments} we show the experimental results. Finally, we conclude and describe future work in Section \ref{Conclusion}.
\section{Related Work} \label{RelatedWork}
In this section, we introduce popular explainability methods, as well as existing approaches for evaluating explanations.

\textbf{Explainability Methods:} Prevailing explainability methods are broadly split into two categories, i.e., gradient-based and perturbation-based. Gradient-based explanations originate from Vanilla Gradients \cite{simonyan2013deep} and demonstrate the importance by calculating the gradient of an output neuron with respect to the input. Improved variants have been proposed for the flaws of the pioneer, e.g., \cite{sundararajan2016gradients} argued that Vanilla Gradients suffers from gradient saturation, thus Integrated Gradients \cite{sundararajan2017axiomatic} is proposed, which computes the integration of gradients starting from a selected uninformative baseline. \cite{smilkov2017smoothgrad} smooths the appearance of the explanations by introducing Gaussian noise to adjacent pixels. \cite{bach2015pixel} proposed Layer-wise Relevance Propagation (LRP) that follows specific rules to propagate importance backwards from the output layer to the input. In addition, there are studies that employ the trick for visually optimizing the explanations, such as \cite{springenberg2014striving,selvaraju2017grad}. Perturbation-based approaches mostly utilize surrogate models that substitute the original black box model with an interpretable linear one, and sample training data locally in the neighborhood of the instance to be explained so that the performance of the surrogate model is similar to that of the original one. Representatives include LIME \cite{ribeiro2016should} and KernelSHAP \cite{lundberg2017unified}, which differ in sampling weights and perturbation patterns. Subsequently, numerous variants are proposed to refine the quality of the explanations, such as \cite{ribeiro2018anchors} and \cite{petsiuk2018rise}. In addition, instance-based explainability methods are widely adopted as well, such as counterfactuals \cite{verma2020counterfactual} and activation maximization \cite{erhan2009visualizing}.

\textbf{Evaluation Approaches:} There are two existing mainstream evaluation methods, the sensitivities to input perturbations and to model parameters \cite{vilone2021notions}. The former operates by perturbing the features of the input and observing the changes in the prediction. For a plausible explanation, perturbations to the feature with the largest attribution lead to severe corruption in the prediction confidence. This approach is intuitive and therefore widely applied to justify the reliability of explanations \cite{bach2015pixel,alvarez2018robustness,kindermans2019reliability,ancona2017towards,ghorbani2019interpretation,samek2016evaluating}. However, it has been scepticized that such perturbations neglect the correlation between features (pixels) or feed a distribution that the model has never learned before, which impairs the reliability of the evaluation \cite{sturmfels2020visualizing,hooker2019benchmark}. \cite{hooker2019benchmark} proposes RemOve And Retrain (ROAR), which retrains the model with the perturbed dataset and observes the magnitude of the performance degradation, to some extent eliminating the out-of-distribution issue of the perturbed data. 

The sensitivity of the model parameters originates from a concern about the sanity of the explanations. \cite{adebayo2018local,adebayo2018sanity} reveal that by randomizing the parameters of the model, several explainability methods remain unaffected, which raises questions about whether the explanations are faithful to the model.

Besides, several other approaches are proposed, such as Pointing Game \cite{zhang2018top}, which counts the number of times the point with the largest attribution in the explanation is inside the target object in the image. User study is also an important methodology to assess explanations. Although the explanation needs to be human-friendly, however, this method is costly and subjective, which may not reveal the true basis of the prediction \cite{petsiuk2018rise}. Moreover, several studies raise concerns about the robustness or stability of explainability methods \cite{alvarez2018robustness,ajalloeian2022smoothed,kindermans2019reliability}. Although they experimentally demonstrate certain deficiencies of explanations, such as the lack of linear invariance \cite{kindermans2019reliability} and Lipschitz continuity \cite{alvarez2018robustness}, they cannot be used as quantitative evaluation approaches.

\section{Methods} \label{Methods}
%Existing approaches for explanation assessment can be broadly classified as instance-based and model-based. Existing explanation assessments fall into two broad categories, namely, input-faithful and model-faithful. The former focuses only on whether the explanation is semantically correlated with the input, e.g., user studies \cite{narayanan2018humans,alqaraawi2020evaluating} and the Pointing Game \cite{zhang2018top}, while the latter focuses only on the relevance of the explanation to the model, e.g., the Sensitivity Perturbation \cite{bach2015pixel}. The rest of the methods are even independent of the model to be explained, such as ROAR \cite{hooker2019benchmark} (they retrain new models for each perturbation). 
In this section we present a novel approach to evaluating explanations. Unlike existing intuitive perturbation-based sensitivity measures, this method targets the generalizability of the explanations across the entire data set. 

Consider a image dataset $X=\left \{x_1,...,x_n \right \} \subseteq R^{w\times h}$ with a well-trained model $H(\cdot)$. We obtain the explanation set $P=\left \{ p_1,...,p_n \right \} \subseteq R^{w\times h}$ by the explainability method $F$. The explainability approach can be regarded as a mapping function, i.e. $p_i=F(H,x_i)$ (only local methods are considered here). Therefore, there should be a certain degree of distributional similarity between $X$ and $P$: $X\mathrel{\dot\sim}P$. Intuitively, this can be understood as "proximity", i.e., similar inputs should yield similar explanations and vise versa. Existing studies had mentioned similar concerns, they perturb the input and measured the local continuity of the explanations with \textit{Local Lipschitz Continuity}, which is formulated as:
\begin{equation}
    \left \| F(H,x_i)-F(H,(x_i+\epsilon) \right \|\leq L\left \| x_i-(x_i+\epsilon ) \right \|
\end{equation}
where $\epsilon$ is the perturbation matrix and $L$ is a constant. However, we argue that the perturbations may disrupt the data distribution. Analogous to sensitivity tests \cite{hooker2019benchmark}, out-of-distribution data that are never seen by the model may may impair the evaluation performance.

Our approach involves the entire process of explanation generation. For a well-trained model $H$, a certain rule $R_m$ should exist with regard to the prediction, for instance, when multiple horse images are input, the model takes similar features as prediction bases(e.g., legs or tails). For a plausible explainability method $F$, it should exhibit the rule in the explanations. However, $R_m$ is agnostic due to the opaqueness of the black box model. Therefore, under the premise that the model is well-trained, we can verify whether $F$ is reliable by verifying $R_m$. Inspired by early researches, which have shown that a network with sufficient parameters can fit any function \cite{hornik1991approximation,cybenko1989approximation}, we train an appropriately structured generative neural network $G$ to simulate $R_m$. $G$ takes the original images $x_i$ as input and outputs the corresponding explanation $p'_i$ ($p'_i  \approx p_i=F(H,x_i)$), and the reliability of the explainability method $F$ is evaluated by observing the performance of $G(X)$ on the whole dataset. The evaluation via G is based on two factors:
\begin{itemize}
    \item Generalizability: $F$ should not be erratic. If the performance of the trained G is inferior, it indicates that $F$ is \emph{unlearnable} and thus \emph{erratic}. For example, $G$ cannot learn any rule from randomly distributed saliency maps, which leads to a nearly non-decreasing loss curve. Note that the reverse of this factor does not hold. One trick is to generate simple and identical explanations for all inputs, which can be easily learned by the network, whereas they cannot be considered as reliable explanations.
    \item Distributional proximity: As a complement to the previous point, $P'$ should possess similar statistical properties as $X$. Intuitively, the gap between explanations in the same class should be much smaller than that between different classes. This complement prevents the aforementioned "simple and identical" explanations from being regarded as plausible, since they are not consistent with the distribution of the original images.
\end{itemize}

\begin{figure}
    \begin{centering}
    \includegraphics[width=0.475\textwidth]{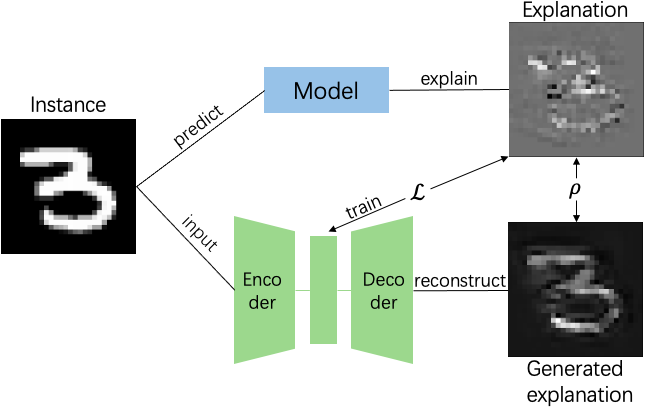}
    \caption{An overview of the evaluation methods. We first select the explainability method to be evaluated and generate explanations utilizing the classification model and the original inputs. Subsequently, we train a generative model that takes the original image as input and attempts a reconstruction of the generated explanations. Finally, we compare the distributional relationships between the reconstructed instances and the explanations.}
    \label{Fig:overview}
    \end{centering}
\end{figure}

Subsequently, we elaborate on the details of the proposed evaluation approach. The general structure of the approach can be seen in Fig. \ref{Fig:overview}. Our evaluation method consists of two components, which verify generalizability and proximity, respectively.

\textbf{Distribution Learnability}: The essential task of the simulatin g model is to learn the rules of the explainability method. Therefore, the model requires sufficient learning capability (i.e., structural complexity). We try an Autoencoder with different architectures until it can reconstruct the input image with high quality ($L_1$ loss $\approx 0.01$, $L_2$ loss $\approx 0.001$ and Structural Similarity Index Measure (SSIM)$>0.99$, the detailed architecture can be seen in Fig. \ref{fig:AE_structure} and the qualitative results are shown Fig. \ref{fig:AE_performance}).  We train this autoencoder with the original images and the explanations as inputs and labels, respectively, with $L1$ as the target loss function until the loss curve converges. We denote this Autoencoder as $AE_i$ for easy reuse in subsequent processes. We divide the dataset $X$ into training ($X_{tr}$) and test sets ($X_{te}$), and we train the Autoencoder corresponding to each explainability method (denoted as $AE_F$, where $F$ represents a certain explainability method) with $X_{tr}$ according to the framework in Fig. \ref{fig:AE_structure}, and then input the $X_{te}$ into $AE_F$ to obtain the reconstructed explanations $P'_{te}$. According to the reconstruction performance the plausibility of the explanations can be observed: Autoencoder is more likely to reconstruct those explanations that possess more typical rules, while on the contrary, erratic and random explanations cannot be well-learned. We show an example for the generalizability evaluation in Section \ref{sec:eval_example}.

To quantitatively assess the learnability of the explanations, we calculate the reconstruction performance of $P'_{te}$ by six different measurements:

\emph{L1\&L2}: The L1 and L2 distances provide an intuitive indication of the pixel-wise similarity, however they may be affected by extreme values and those points with little attribution occupy the same weight.

\emph{Pearson \& Spearman's rank correlation coefficient}: Pearson correlation coefficient (PC) measures the linear correlation of two explanations, which is formulated as $\rho =\frac{COV(P,P')}{\sigma_{P}\sigma_{P'}}$. When comparing explanations, one may focus more on the ranking of feature attributions than on the specific values. Spearman's correlation coefficient (SC) is the ranked version of the \emph{Pearson correlation coefficient}, which can be formulated as:
\begin{equation}
    \rho_R =\frac{COV(R(P),R(P'))}{\sigma_{R(P)}\sigma_{R(P')}}
\end{equation}
where $R(*)$ is the rank function, COV is the covariance and $\rho$ is standard deviation. 

\emph{Top-k Accuracy (TA)}: For explanations, humans tend to be concerned only with the features that yield larger attributions. Therefore, we perform a Pointing Game on the pixels with the top-K attributions. Specifying a percentage K (all $25\%$ in the experiment), we calculate how many pixels in $P_{te}$ ranked in the top-K attributions appear in the top-K attributions of the pixels in $P'_{te}$. It can be formulated as:
\begin{equation}
    TA=\frac{\left | TopK(P)\cap TopK(P') \right |}{\left | TopK(P') \right |}
\end{equation}
TA is in the range $[0,1]$, with higher values representing that the pixels with large attribution are more similar in the two explanations. Note that the TA of two sufficiently large random sequences converge to $K$ (0.25 in this experiment).

\emph{Structural Similarity Index Measure (SSIM)}: In particular, for images, we introduce SSIM as a similarity measure, which measures the statistical likelihood of two explanations regarding the mean and variance in terms of luminance, contrast and structure. SSIM is formulated as:
\begin{equation}
    SSIM(P,P')=\frac{(2\mu_P\mu_{P'}+c_1)(2\sigma_{PP'}+c_2)}{(\mu_P^2+\mu_{P'}^2+c_1)(\sigma_P^2+\sigma_{P'}^2+c_2)}
\end{equation}
where $\mu$ and $\sigma$ denote the mean and (co)variance, respectively. Note that the core of SSIM is the approximation of the statistics, existing studies show that it is problematic to assess the perceptual proximity \cite{nilsson2020understanding}, so we only consider it as a reference.

\textbf{Variance Proximity}: Good generalizability is not sufficient to conclude the plausibility of the explanations. Simple rules can be easily learned by the Autoencoder (e.g. by focusing the attribution on a fixed pixel), however they may not provide reliable explanations. Therefore, we exclude these uninformative explanations by evaluating the proximity of the distributions between the reconstructed explanations and the original images. Our assessment is based on the idea that the discrepancies of the reconstructed explanations between the identical classes should be much smaller than those between different categories. In this regard we consider two proximity measurements, the pixel-wise ranking discrepancy and the latent distance. For the former, calculating the absolute value is insignificant, since the reconstructed explanations may not be in the same order of magnitude as the original image. Therefore, we also employ \textit{Spearman's rank correlation coefficient} to measure pixel-wise proximity. For the latter, we constructed a latent distance measurement for validation with the pixel-wise approach, which can be considered as a variant of \textit{Fréchet inception distance (FID)}. We take the encoder of the original Autoencoder ($AE_i$) as the "Inception" network, input the images and explanations respectively to obtain the latent vectors, and compute their 2-Wasserstein distances (both are considered as multidimensional Gaussian distributions)
\begin{equation}
\begin{aligned}
    &d_F(L_P(\mu_P,\sigma_P),L_{P'}(\mu_{P'},\sigma_{P'}))= \left \| \mu_P-\mu_{P'} \right \|_{2}^{2} \\
    & +tr\left ( \sigma_P+\sigma_{P'} -2\left ( \sigma_P^{\frac{1}{2}}\cdot \sigma_{P'}\cdot\sigma_P^{\frac{1}{2}} \right )^\frac{1}{2} \right )
\end{aligned}
\end{equation}

In measuring the reconstructed similarity within classes, we take a certain number of samples ($S$) from each class for a pairwise comparison. For inter-class, we try all combinations of different classes and take $S$ samples from each class for a pairwise comparison. For quantitative comparison, our final Variance Proximity scores (VP) can be calculated as:
\begin{equation} \label{eq:VPscore}
    VP=\Delta{\overline{SC_a^r}}+\Delta{\left \|\overline{FID_a^r}\right \|}
\end{equation}
where $\Delta{\overline{SC_a^r}}$ is the difference between the mean of the inter- and intra-class Spearman coefficients, which can be formulated as:
\begin{equation}
    \Delta{\overline{SC_a^r}} = \overline{SC_{a}} -  \overline{SC_{r}}
\end{equation}
and $\Delta{\left \|\overline{FID_a^r}\right \|}$ denotes the (normalized) difference between the averages of the inter- and intra-class FIDs, which can be expressed as:
\begin{equation}
    \Delta{\left \|\overline{FID_a^r}\right \|}=(\overline{FID_{r}} -  \overline{FID_{a}})/\overline{FID_{a}}
\end{equation}

Finally, the quantitative score of corresponding explainability methods is:
\begin{equation}
    S_{EM}=VP\times DL
\end{equation}

where VP is the variance proximity (see equation \ref{eq:VPscore}), and DL denotes the Distribution Learnability, which is the average of Top-k Accuracy (TA), Spearman's Correlation (SC) and Pearson Correlation (PC) coefficients.
\section{Experiments} \label{Experiments}
In this section, we first show an example comparison on evaluating LIME (Section \ref{sec:eval_example}), followed by a quantitative evaluation of the popular explainability methods with the proposed approach. We choose MNIST handwritten dataset for experiments. The accuracy of the classification model to be explained is $98.5\%$ on the test set. The structure of Autoencoders is identical, whose latent vector has dimension 128. We train each for $100$ epochs with a learning rate of $1e-5$. The Autoencoder that reconstructs the original image achieves $L1 loss \approx 0.01$, $MSE loss \approx 0.001$ and $SSIM > 0.99$. In the evaluation of Variance Proximity, we choose $S=500$.

\subsection{Case study:The Generalizability of LIME} \label{sec:eval_example}
To intuitively demonstrate the correlation between the generalizability and the quality of explanations, we first show a straightforward evaluation example. We choose an explainability approach by tuning the parameters to generate explanations with different qualities. LIME \cite{ribeiro2016should} is an ideal candidate because it relies on perturbing the instances to be explained to try out local decision boundaries, and the number of perturbed samples affects the quality of the explanation. The prior knowledge already exists that LIME performs poorly with few perturbed samples and improves as they increase until saturation. We segment the input into 50 super-pixels, and choose LIME with a total number of perturbed samples ($n\_samples$) of 10, 30, 50, 100 and 500 for explaining the classification model respectively, and evaluate the generalizability of the explanations according to the proposed method. 

\begin{figure*}
    \begin{centering}
    \includegraphics[width=0.8\textwidth]{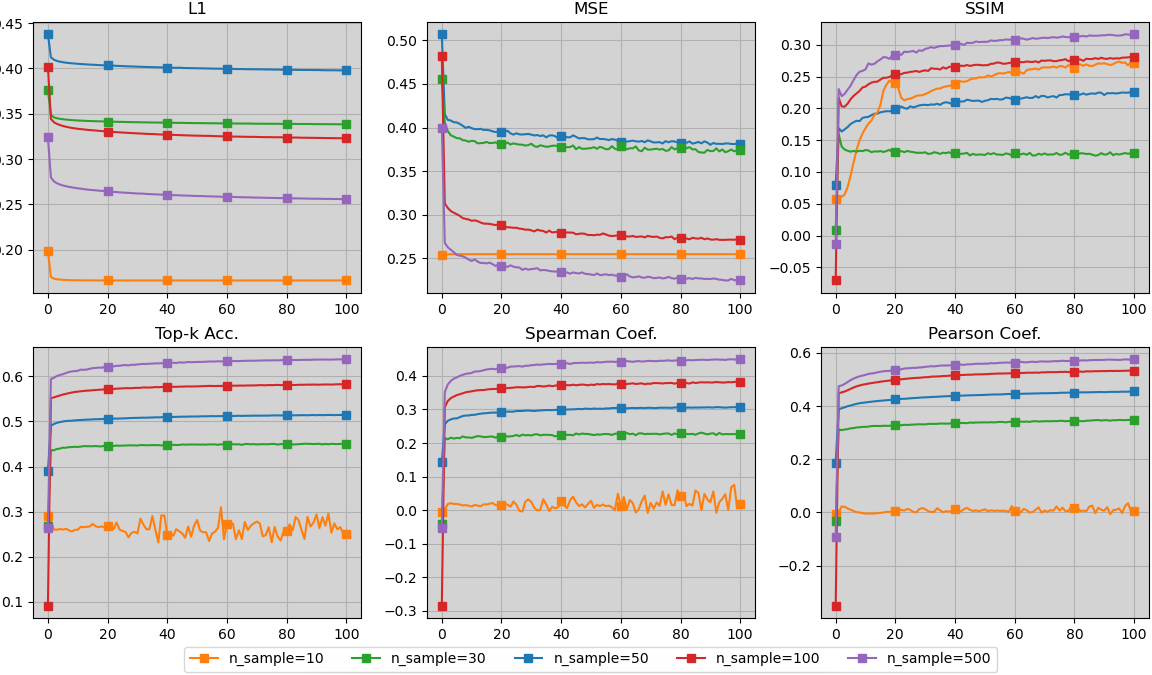}
    \caption{The training curves of LIME with perturbed sample numbers of 10 (yellow), 30 (green), 50 (blue), 100 (red), and 500 (purple), respectively. The y-axis is the value of the corresponding metrics and the x-axis is the training epoch numbers.}
    \label{fig:LIME_compare}
    \end{centering}
\end{figure*}

The results, as illustrated in Fig. \ref{fig:LIME_compare}, indicate that insufficient number of perturbation samples may prevent Autoencoder from learning the distribution of the explanations. For instance, during training, although the loss in value (L1,MSE and SSIM) of LIME10 ($n\_sample=10$) is at a low level ($0.165\pm 1e-4$, $0.254\pm 1e-5$ and $0.271\pm 0.004$, respectively, the last $10$ epochs are statistically involved), the distribution is barely learned: the pixel accuracy of top-k ($\overline{TA}$) converges to $25\%$ ($26 \pm 3.6\%$), which is almost close to the random distribution, and both Spearman and Pearson coefficients (SC and PC) oscillate near zero ($0.021 \pm 0.054$ and $0.020 \pm 0.056$, respectively), implying there is hardly (ranked) linear correlation between the learned distribution and original explanation. As $n\_sample$ increases, the distribution learned by Autoencoder from the explanations grows more accurate. It can be observed that when $n\_sample$ increases to $30$, the top-k accuracy, Spearman and Pearson coefficients of the generated samples converge to $44.9 \pm 0.1\%$, $0.227 \pm 0.003$ and $0.347 \pm 0.002$, respectively. As $n\_sample$ further grows to $50$, these metrics are raised to $51.4 \pm 0.1\%$, $0.306 \pm 0.002$ and $0.454 \pm 0.003$, respectively. However, the benefits from a continued increase in $n\_sample$ are limited ($\overline{TA}$, $\overline{SC}$ and $\overline{PC}$ converge to $63.7 \pm 0.1\%$, $0.448 \pm 0.001$ and $0.574 \pm 0.002$, respectively when $n\_sample=500$). We report the final DL scores of all LIME explanations as $0.103$, $0.341$, $0.425$, $0.498$ and $0.553$, respectively.

\begin{figure*}
    \begin{centering}
    \includegraphics[width=1\textwidth]{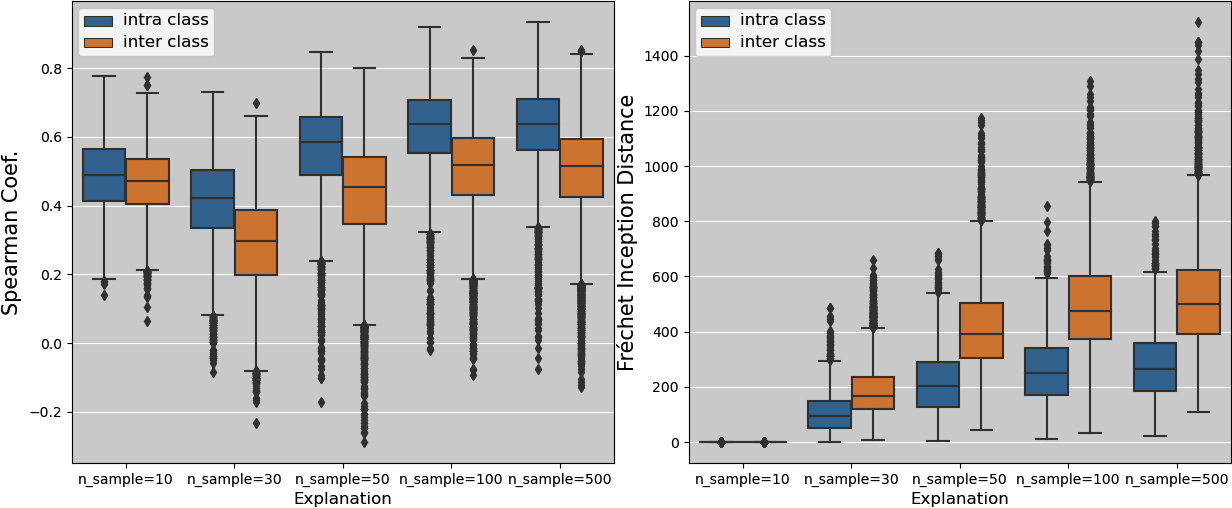}
    \caption{The intra (blue) and inter (orange) class similarity (discrepancy) of the samples generated by Autoencoder based on the explanations of LIME with different number of perturbations. The x-axis from left to right shows the LIME for 10, 30, 50, 100 and 500 perturbed samples, respectively, and the y-axis is the Spearman coefficient (left) and Fréchet Inception Distance (right), respectively. Note that large Spearman coefficients represent similar distributions, while FIDs are the opposite.}
    \label{fig:LIME_class}
    \end{centering}
\end{figure*}

To avoid the trap of "simple rules", we also evaluate the variances and proximities between the generated samples. Fig. \ref{fig:LIME_class} depicts the point-wise ranking (left) and latent space (right) proximity of the generated samples for inter and intra-classes. We observe that as the number of perturbed samples grows, the Spearman coefficients (SC) increase in varying degrees according to the class relationship. When the number of perturbation samples is minimal ($n\_sample=10$), the similarity of intra- and inter-classes is approximated. The average of SC for inter and intra-class are $\overline{SC_{r}}=0.46$ and $\overline{SC_{a}}=0.49$, respectively, whose difference is $\Delta{\left |\overline{SC_a^r}  \right |} = 0.02$. Similarly, the means of FID for inter and intra-class are $\overline{FID_{r}}=1.4e-5$ and $\overline{FID_{a}}=1.8e-5$, respectively, the corresponding difference is $\Delta{\left |\overline{FID_a^r}  \right |} = 3.7e-6$). According to the boxplot, it can be observed that the gaps between inter and intra-classes obviously increase with the growth of $n\_sample$. According to equation \ref{eq:VPscore}, we derive the VP scores of LIME with different n\_sample as 0.316, 0.877, 1.017, 1.000 and 0.973, respectively.

We finally report the quantitative scores of LIME with 10, 30, 50, 100 and 500 perturbed samples as 0.031, 0.296, 0.429, 0.498, 0.536. The detailed tabulated results are shown in Table \ref{tab:compareLIME}. For intuitive understanding, we present the explanations generated by LIME with different $n\_sample$ in Fig. \ref{fig:qualitative_view}.

\subsection{The Generalizability of Explainability Methods} \label{sec:eval_all}
A further step is to extend the proposed evaluation approach to more explainability methods. We choose several popular gradient-based and perturbation-based methods, including Vanilla Gradients \cite{simonyan2013deep}, Guided Back-propagation \cite{springenberg2014striving}, Input$\times$Gradients \cite{shrikumar2016not}, Integrated Gradients \cite{sundararajan2017axiomatic}, Layer-wise Relevance Propagation \cite{bach2015pixel}, DeepLift \cite{shrikumar2016not}, LIME ($n_sample=100$) \cite{ribeiro2016should} and KernelSHAP \cite{lundberg2017unified}. As a reference, we additionally introduce a randomly generated noise explanation. We again make predictions and explanations on MNIST for all test sets and reconstruct the explanations with the Autoencoder of the same architecture. 

\begin{figure*}
    \begin{centering}
    \includegraphics[width=0.8\textwidth]{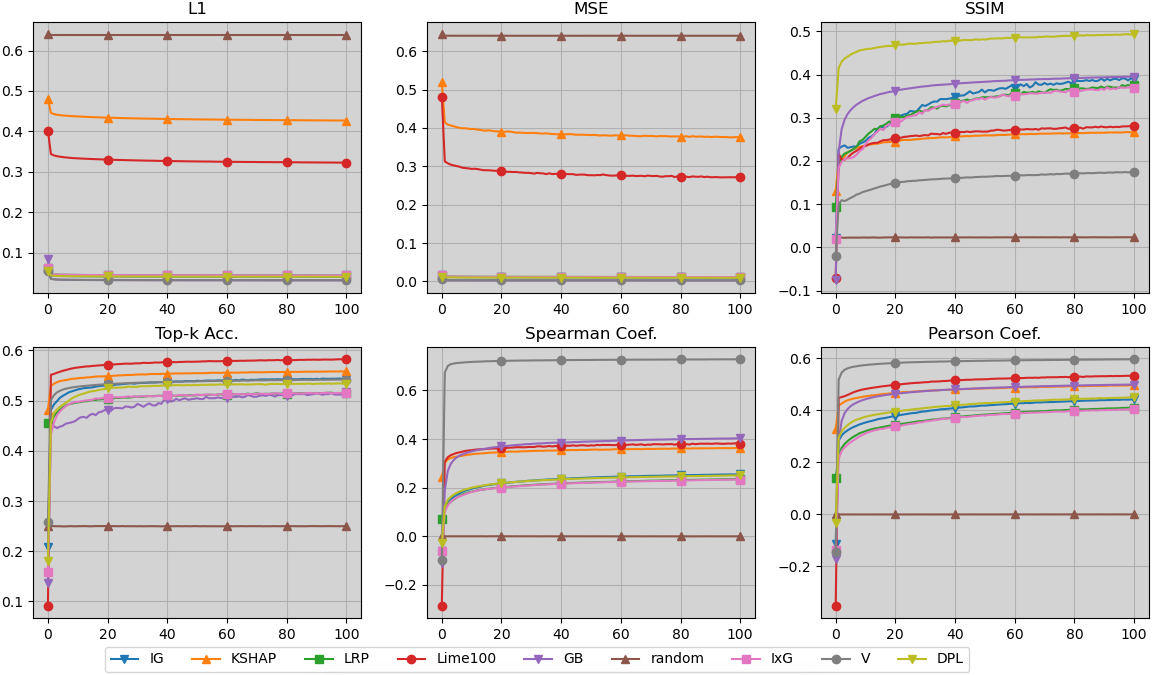}
    \caption{The training curves of Vanilla Gradients, GB, IxG, IG, LRP, DeepLift, LIME, KernelSHAP and random explanation, respectively. The y-axis is the value of the corresponding metrics and the x-axis is the training epoch numbers.}
    \label{fig:compare_all}
    \end{centering}
\end{figure*}

The training results are illustrated in Fig. \ref{fig:compare_all}. In terms of loss (above row), all methods (except random) can reduce the distance and statistical gap (L1, MSE and SSIM), which implies that all Explainability methods possess a degree of generalizability. This is also evidenced from the perspective of the proximity of the distribution (bottom row): All of the explainability methods far outperformed the random explanation in the three metrics, namely Top-K accuracy, Spearman and Pearson coefficients. Among these approaches, Vanilla Gradients ($\overline{TA}=0.541\pm2e-4$, $\overline{SC}=0.727\pm2e-4$ and $\overline{PC}=0.596\pm4e-4$, calculated by averaging the last 10 epochs), LIME ($\overline{TA}=0.582\pm5e-4$, $\overline{SC}=0.381\pm1e-3$ and $\overline{PC}=0.532\pm8e-4$) and KernelSHAP ($\overline{TA}=0.558\pm4e-4$, $\overline{SC}=0.362\pm8e-4$ and $\overline{PC}=0.496\pm1e-3$) slightly outperform in terms of learnability of the distribution. Comparatively, IxG and LRP are less learnable, with all three metrics being relatively inferior ($\overline{TA}=0.515\pm8e-4$, $\overline{SC}=0.232\pm8e-4$,
$\overline{PC}=0.403\pm1e-3$ and $\overline{TA}=0.515\pm7e-4$, $\overline{SC}=0.234\pm9e-4$,
$\overline{PC}=0.409\pm1e-3$, respectively). Furthermore, we observe that the generalizability of perturbation-based methods is superior to that of the majority of gradient-based methods ($\overline{TA_{overall}}=0.570$, $\overline{SC_{overall}}=0.371$) and $\overline{PC_{overall}}=0.514$. 

\begin{figure*}
    \begin{centering}
    \includegraphics[width=1\textwidth]{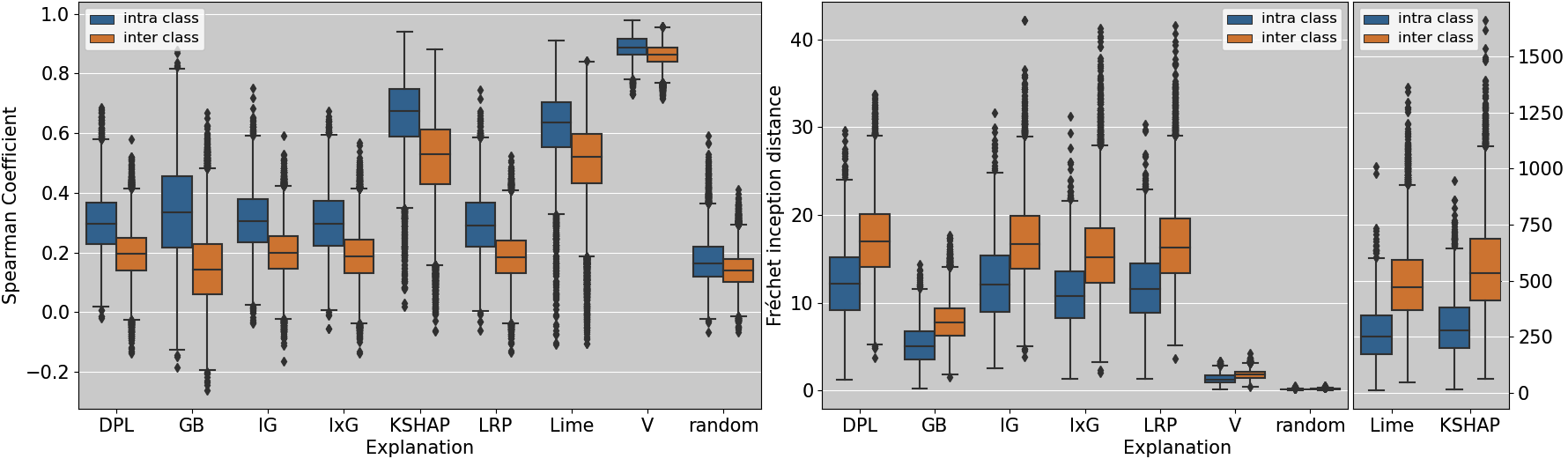}
    \caption{The intra (blue) and inter (orange) class similarity (discrepancy) of the samples generated by Autoencoder based on the explanations of various explainability approaches. DPL, GB, IG IxG, KSHAP, and V denote DeepLift, Guided Backpropagation, Integrated Gradients, Input$\times$Gradients, KernelSHAP, and Vanilla Gradients, respectively, and the y-axis is the Spearman coefficient (left) and Fréchet Inception Distance (right), respectively. The FIDs of the perturbation-based explanations are separated since they are not in the same order of magnitude as the rest.}
    \label{fig:all_class}
    \end{centering}
\end{figure*}

Again, we assess the Variance Proximity between the generated explanations. As demonstrated in Fig. \ref{fig:all_class}, the reconstructed explanations of all explainability methods outperform the random ones, which represents that their explanations are learnable in terms of intra-class and inter-class discrepancies. However, Vanilla Gradients ($\Delta\overline{SC_a^r}=0.025$, $\Delta\overline{FID_a^r}=0.364$ and $VP=0.389$), Integrated Gradients ($\Delta\overline{SC_a^r}=0.105$, $\Delta\overline{FID_a^r}=0.386$ and $VP=0.491$) and DeepLift ($\Delta\overline{SC_a^r}=0.107$, $\Delta\overline{FID_a^r}=0.378$ and $VP=0.485$) are relatively inferior. Considering the excellent performance of Vanilla Gradients ($DL=0.622$) in Distribution Learnability, we believe that the explanation distributions are relatively homogeneous, in line with the aforementioned "simple rule" trap. IG and DeepLift perform mediocrely in Distributional Learnability ($DL=0.413$ and $0.411$, respectively), and therefore we consider their explanations as slightly noisy, which interfere with the learning of the Autoencoder. In addition, we observe that the perturbation-based approaches perform better in the assessment of Variance Proximity ($VP=0.963$ for LIME and $1.016$ for KernelSHAP). This is mainly attributed to the randomness and complexity of the perturbation-based explanation generating process, which prevents Autoencoder from summarizing fixed and simplistic rules.

As a conclusion, we report that in comparison to gradient-based methods ($S_{EM}$ for $V$, $GB$, $I\times G$, $IG$, $LRP$ and $DPL$ are $0.241$, $0.330$, $0.196$, $0.200$, $0.203$ and $0.201$, respectively), perturbation-based ones show better performance ($S_{EM}$ for $LIME$ and $KSHAP$ are both $0.479$) since the distributions of their generated explanations are more generalizable. For a clearer comparison, tabulated results for all metrics and the final quantitative scores are presented in Table \ref{tab:compareall}. Again, we show in Fig. \ref{fig:qualitative_all} a comparison of several visualizations of the explanations.

\subsection{Smoothed vs. Unsmoothed Maps} \label{sec:eval_smooth}
SmoothGrad is a technique that finds the gradient (or other method to obtain the explanations) after introducing noise to the original image multiple times and takes the average value. SmoothGrad-processed explanations are considered cleaner and more comprehensible. Interestingly, we found that SmoothGrad \cite{smilkov2017smoothgrad} not only enhances visual consciousness, but also increases the generalizability of the explanations. We choose three gradient-based methods as baselines, namely Vanilla Gradients (V), Input$\times$Gradients (IxG) and Integrated Gradients (IG), and implement SmoothGrad for them respectively. 

\begin{figure*}
    \begin{centering}
    \includegraphics[width=0.8\textwidth]{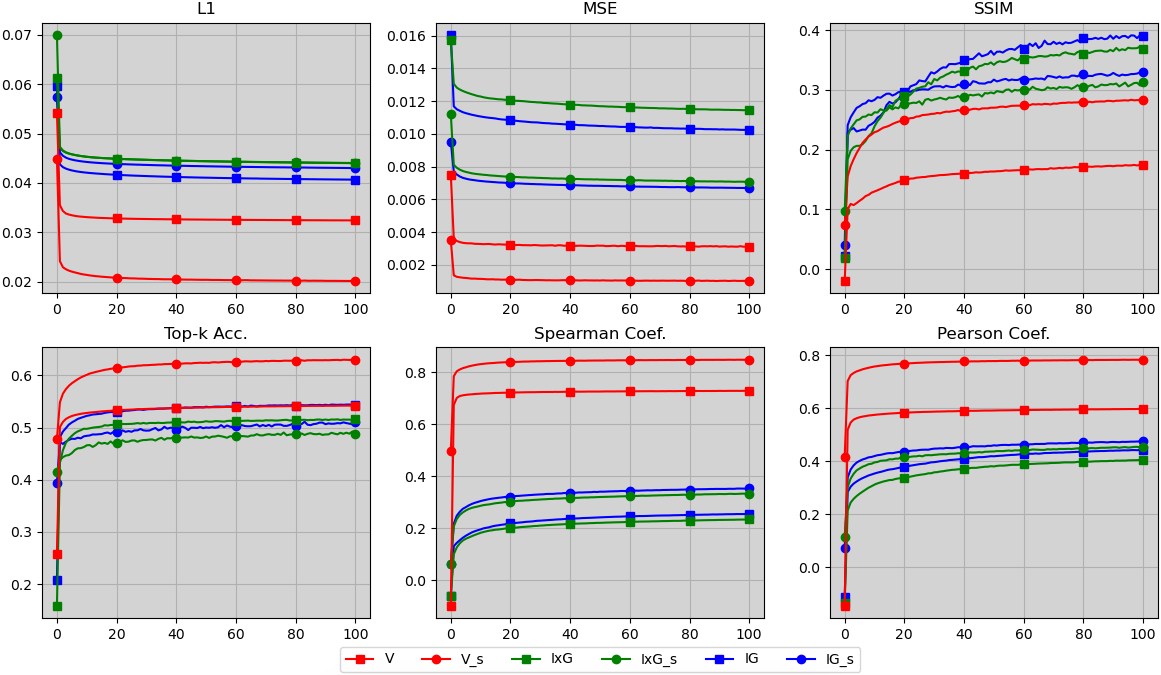}
    \caption{The training curves of Vanilla Gradients (V), Input$times$Gradients (IxG), Integrated Gradients (IG) and their corresponding SmoothGrad versions (with suffix "\_s"), respectively. The y-axis is the value of the corresponding metrics and the x-axis is the training epoch numbers.}
    \label{fig:compare_smooth}
    \end{centering}
\end{figure*}

\begin{figure*}
    \begin{centering}
    \includegraphics[width=1\textwidth,height=6cm]{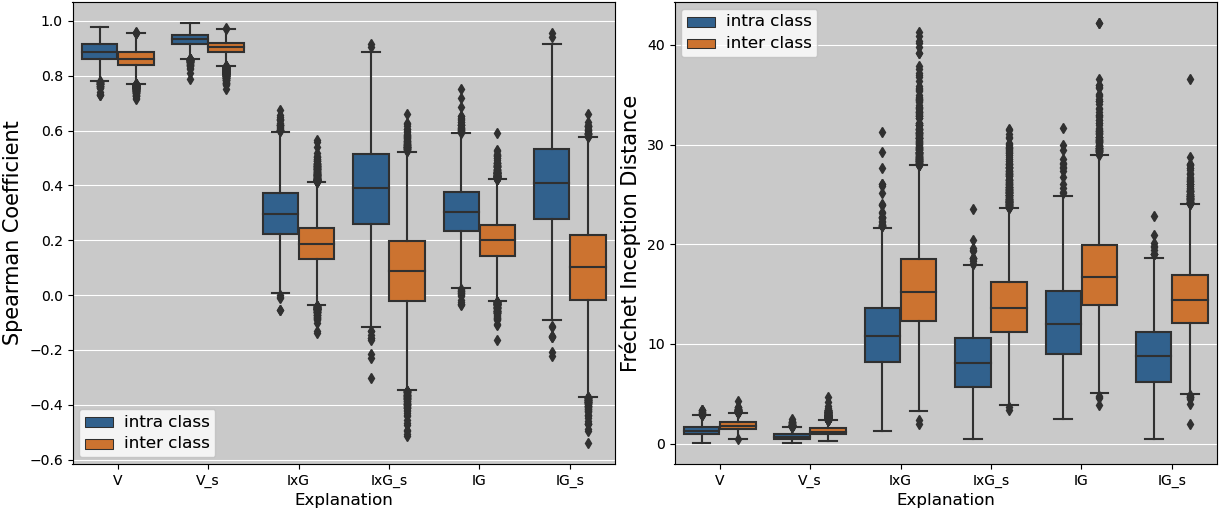}
    \caption{The intra (blue) and inter (orange) class similarity (discrepancy) of the samples generated by Autoencoder based on the explanations of Vanilla Gradients (V), Input$times$Gradients (IxG), Integrated Gradients (IG) and their corresponding SmoothGrad versions (with suffix "\_s"), respectively, and the y-axis is the Spearman coefficient (left) and Fréchet Inception Distance (right), respectively.}
    \label{fig:smooth_class}
    \end{centering}
\end{figure*}

As shown in Fig. \ref{fig:compare_smooth}, for the Distribution Learnability (DL), the SmoothGrad-applied versions all outperform the original baselines (The differences in DL $\Delta{DL}$ are $0.130$, $0.041$ and $0.031$ for V, IxG and IG, respectively), which implies that explanations with SmoothGrad are more learnable and generalizable. On the other hand, as Fig. \ref{fig:smooth_class} illustrated, SmoothGrad significantly strengthens the Variance Proximity (VP) of the explanations. The VPs of the three explainability methods increase by $0.295$, $0.468$ and $0.475$ respectively. As a consequence, SmoothGrad raises the final scores of V, IxG, and IG from $0.241$, $0.196$, and $0.200$ to $0.514$, $0.415$, and $0.426$, respectively, which is consistent with human intuition: \cite{smilkov2017smoothgrad} concludes that SmoothGrad significantly reduces the noise in the saliency maps and thus augments their comprehensibility. We present the detailed quantitative results in Table \ref{tab:smooth} and the two learned explanation distributions in Fig. \ref{fig:qualitative_smooth}.

\section{Conclusion} \label{Conclusion}
This work provides a novel perspective for quantitatively evaluating explainability methods: generalizability. We argue that the distributions of good explanations should have clearer regularities and learnabilities, and possess distributional approximations and variations within and between classes. We demonstrate the evaluation of multiple explainability methods with the proposed approach. In the future work, we look forward to refining quantitative evaluation approaches that are more comprehensive and objective for explainability methods.

\clearpage
{\small
\bibliographystyle{IEEEtran}
\bibliography{egbib}
}

\clearpage
\section{Supplementary Material}
\beginsupplement

In this section,We provide additional materials for the main part of the paper.

\subsection{Architecture \& Performance of the Autoencoder}
\begin{figure*}
    \begin{centering}
    \includegraphics[width=0.6\textwidth]{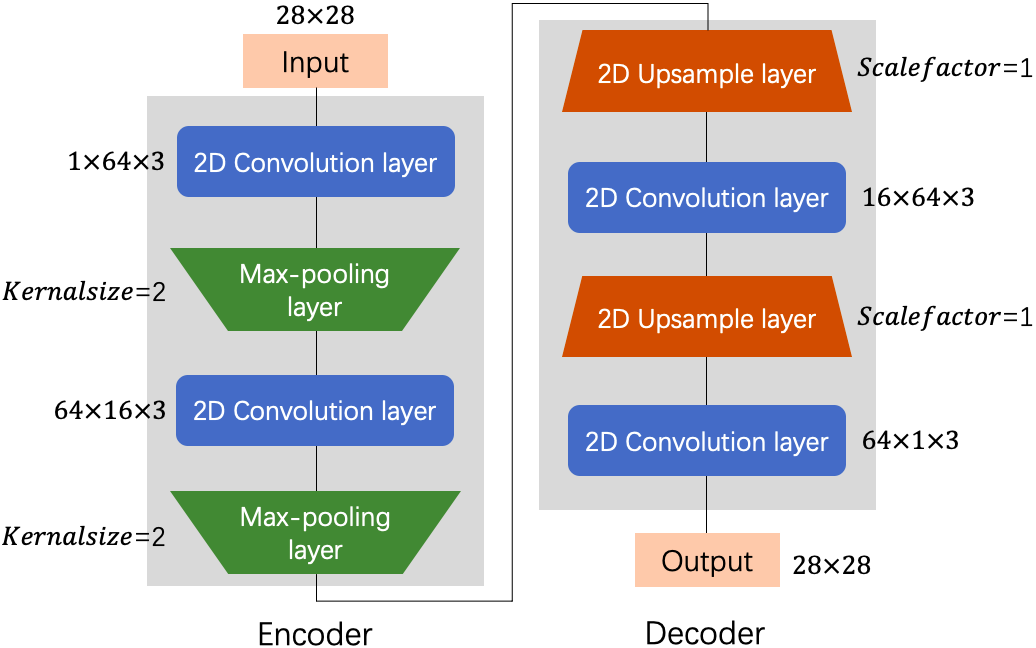}
    \caption{The structure of Autoencoder used to reconstruct the explanations.}
    \label{fig:AE_structure}
    \end{centering}
\end{figure*}

We reconstruct the explanations with a simply structured Autoencoder, which is illustrated in Fig. \ref{fig:AE_structure}.

\begin{figure*}
    \begin{centering}
    \includegraphics[width=0.6\textwidth]{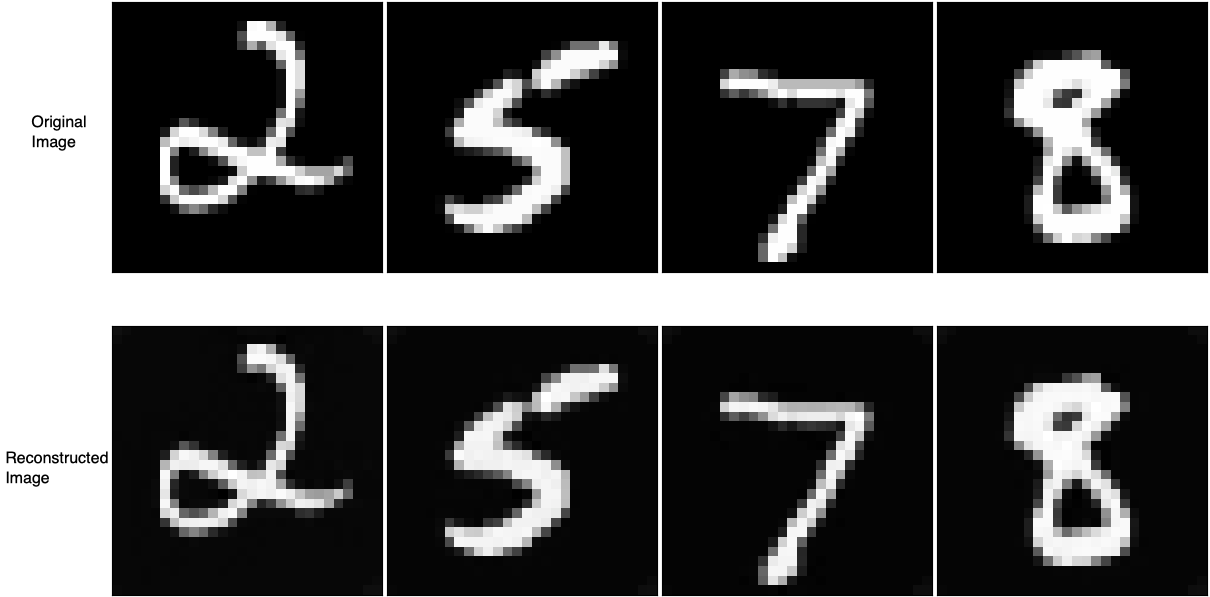}
    \caption{The performance of the Autoencoder reconstructing the MNIST handwriting dataset. The first and second rows are the original images and the reconstructed ones, respectively.}
    \label{fig:AE_performance}
    \end{centering}
\end{figure*}

The qualitative performance of the reconstruction on the original MNIST dataset with the Autoencoder is illustrated in Fig. \ref{fig:AE_performance}. Noticeably, the Autoencoder with this architecture is capable of reconstructing the original image with a high degree of restoration, which indicates that the learning ability is sufficient for the dataset.

\subsection{Supplementary results for Section \ref{sec:eval_example}}

\begin{figure*}
    \begin{centering}
    \includegraphics[width=0.8\textwidth]{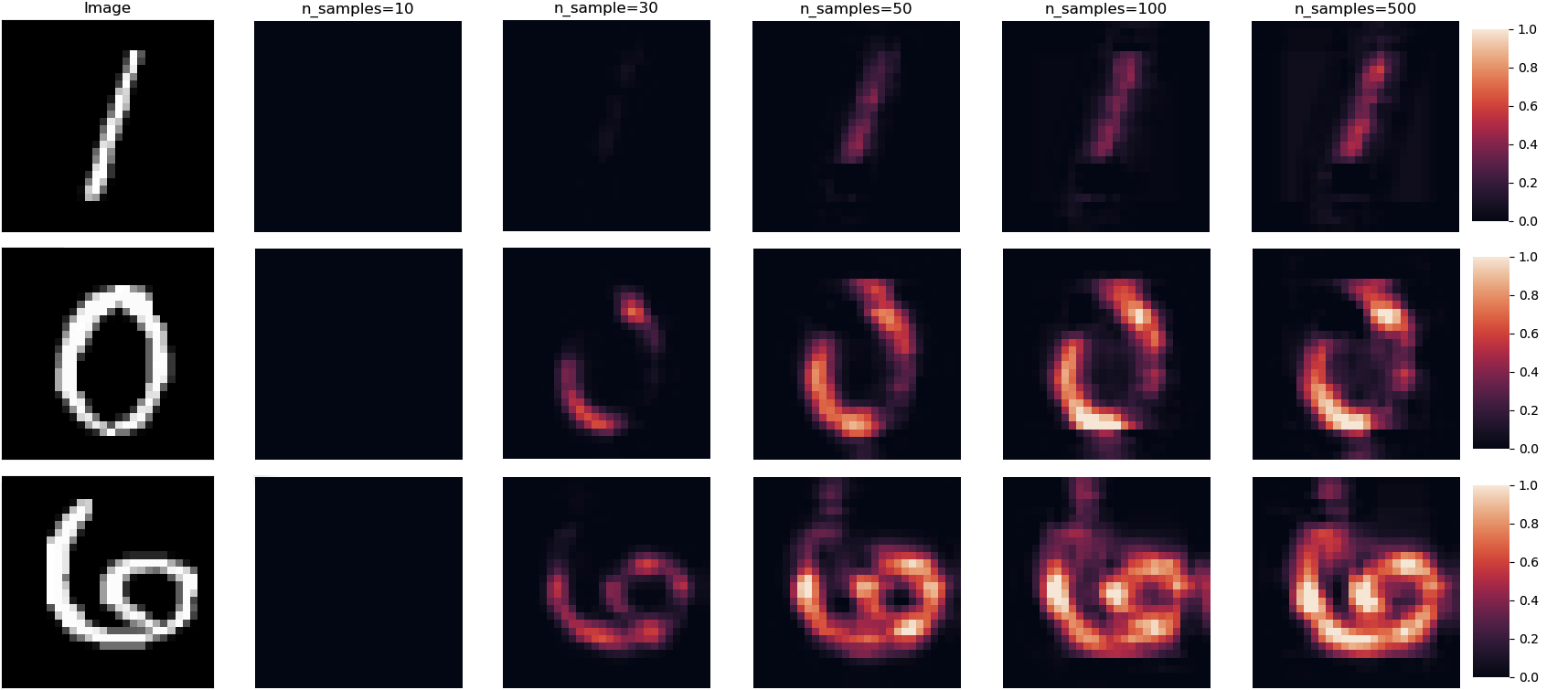}
    \caption{Visualization of the three original images and their corresponding reconstructed explanations learned from LIME with $n\_sample=10$, $30$, $50$, $100$ and $500$, respectively. Note that our approach is based on the evaluation of the statistics on reconstructions and \textit{the clarity of the reconstructed explanations does not directly indicate the quality of the explanations}. For clearer visual effects, please increase the monitor brightness when viewing.}
    \label{fig:qualitative_view}
    \end{centering}
\end{figure*}

In this subsection, we present the detailed results (all metrics are included) for the Case Study in Table \ref{tab:compareLIME}, i.e., the explanation learnability for LIME with different $n\_samples$. In Fig. \ref{fig:qualitative_view}, we illustrate the visualization of three reconstructions of the explanations. Note that the quality or contrast of the reconstructed explanations does not directly indicate the generalizability of the corresponding explanations.

\begin{table}[htp]
\centering
\begin{tabular}{cccccc}
\hline
$n\_sample$                 & 10    & 30    & 50    & 100   & 500   \\ \hline
$\overline{TA}$                       & $26.7$  & $44.9$  & $51.4$  & $58.2$  & $\mathbf{63.7}$  \\
$\overline{SC}$                       & $0.032$ & $0.227$ & $0.306$ & $0.381$ & $\mathbf{0.448}$ \\
$\overline{PC}$                       & $0.010$ & $0.347$ & $0.454$ & $0.532$ & $\mathbf{0.574}$ \\
DL                       & $0.103$ & $0.341$ & $0.425$ & $0.498$ & $\mathbf{0.553}$ \\ \hline
$\Delta\overline{SC_a^r}$  & $0.019$ & $0.120$ & $\mathbf{0.125}$ & $0.111$ & $0.124$ \\
$\Delta\overline{FID_a^r}$ & $0.297$ & $0.757$ & $\mathbf{0.892}$ & $0.889$ & $0.849$ \\
VP                       & $0.316$ & $0.877$ & $\mathbf{1.017}$ & $1.000$ & $0.973$ \\ \hline
$S_{EM}$                   & $0.031$ & $0.296$ & $0.429$ & $0.498$ & $\mathbf{0.536}$ \\ \hline
\end{tabular}
\caption{Detailed quantitative results of the proposed evaluation method for LIME with different $n\_sample$. From top to bottom are the average of: Top-k Accuracy, Spearman's and Pearson's coefficients, the Distribution Learnability, the difference of Spearman's coefficient and Fréchet inception distance between intra and inter-class, the Variance Proximity and the final score of the generalizability.}
\label{tab:compareLIME}
\end{table}

\subsection{Supplementary results for Section \ref{sec:eval_all}}
\begin{figure*}
    \begin{centering}
    \includegraphics[width=0.8\textwidth]{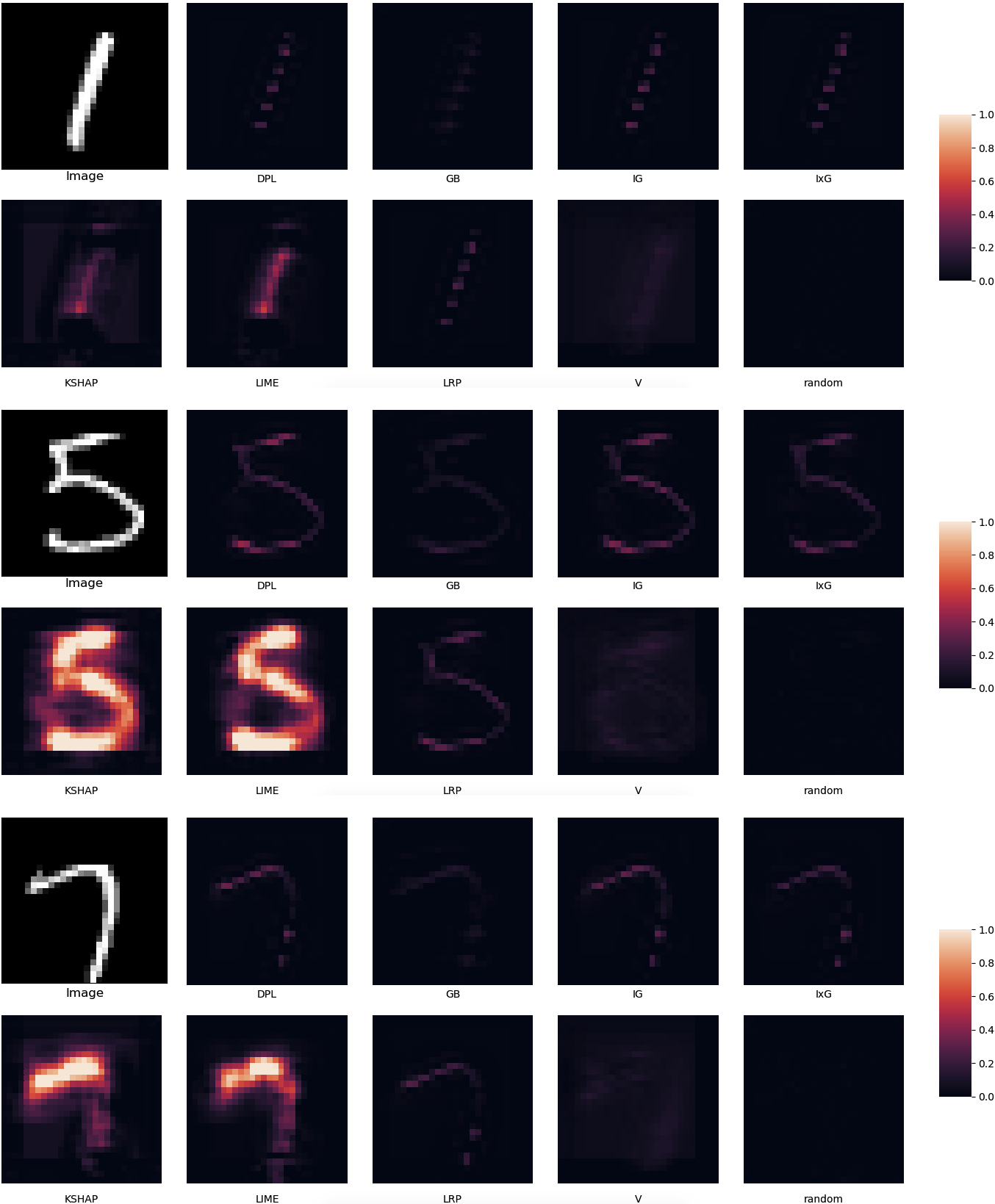}
    \caption{Visualization of the three original images and their corresponding reconstructed explanations learned from popular explainability methods. For clearer visual effects, please increase the monitor brightness when viewing.}
    \label{fig:qualitative_all}
    \end{centering}
\end{figure*}

In this subsection, we first provide detailed tabular results for the evaluation experiments of popular explainability methods in Table \ref{tab:compareall}. Subsequently, we display in Fig. \ref{fig:qualitative_all} three examples of reconstructed explanations by learning the corresponding explainability methods.

\begin{table*}[]
\centering
\begin{tabular}{c|cccccc|cc|c}
\hline
                           & V                & GB               & IxG     & IG      & LRP     & DPL     & LIME             & KSHAP            & Random  \\ \hline
$\overline{TA}$            & $0.541$          & $0.513$          & $0.515$ & $0.543$ & $0.515$ & $0.534$ & $\mathbf{0.582}$ & $0.558$          & $0.250$ \\
$\overline{SC}$            & $\mathbf{0.727}$ & $0.402$          & $0.232$ & $0.254$ & $0.234$ & $0.249$ & $0.381$          & $0.362$          & $-8e- 5$ \\
$\overline{PC}$            & $\mathbf{0.596}$ & $0.499$          & $0.403$ & $0.441$ & $0.409$ & $0.449$ & $0.532$          & $0.496$          & $2e-4$  \\
DL                         & $\mathbf{0.622}$ & $0.471$          & $0.383$ & $0.413$ & $0.386$ & $0.411$ & $0.498$          & $0.472$          & $0.083$ \\ \hline
$\Delta\overline{SC_a^r}$  & $0.025$          & $\mathbf{0.193}$ & $0.111$ & $0.107$ & $0.110$ & $0.105$ & $0.109$          & $0.140$          & $0.035$ \\
$\Delta\overline{FID_a^r}$ & $0.364$          & $0.508$          & $0.401$ & $0.378$ & $0.418$ & $0.386$ & $0.854$          & $\mathbf{0.876}$ & $0.204$ \\
VP                         & $0.389$          & $0.701$          & $0.512$ & $0.485$ & $0.528$ & $0.491$ & $0.963$          & $\mathbf{1.016}$ & $0.239$ \\ \hline
$S_{EM}$                   & $0.241$          & $0.330$          & $0.196$ & $0.200$ & $0.203$ & $0.201$ & $\mathbf{0.479}$ & $\mathbf{0.479}$ & $0.019$ \\ \hline
\end{tabular}
\caption{Detailed quantitative results of the proposed evaluation method for popular explainability methods. From left to right are: Vanilla Gradients, Guided Backpropagation, Input$\times$Gradients, Integrated Gradients, Layer-wise Relevance Propagation, DeepLift, LIME, KernelSHAP and random generated explanations.}
\label{tab:compareall}
\end{table*}

\subsection{Supplementary results for Section \ref{sec:eval_smooth}}

\begin{figure*}
    \begin{centering}
    \includegraphics[width=0.6\textwidth]{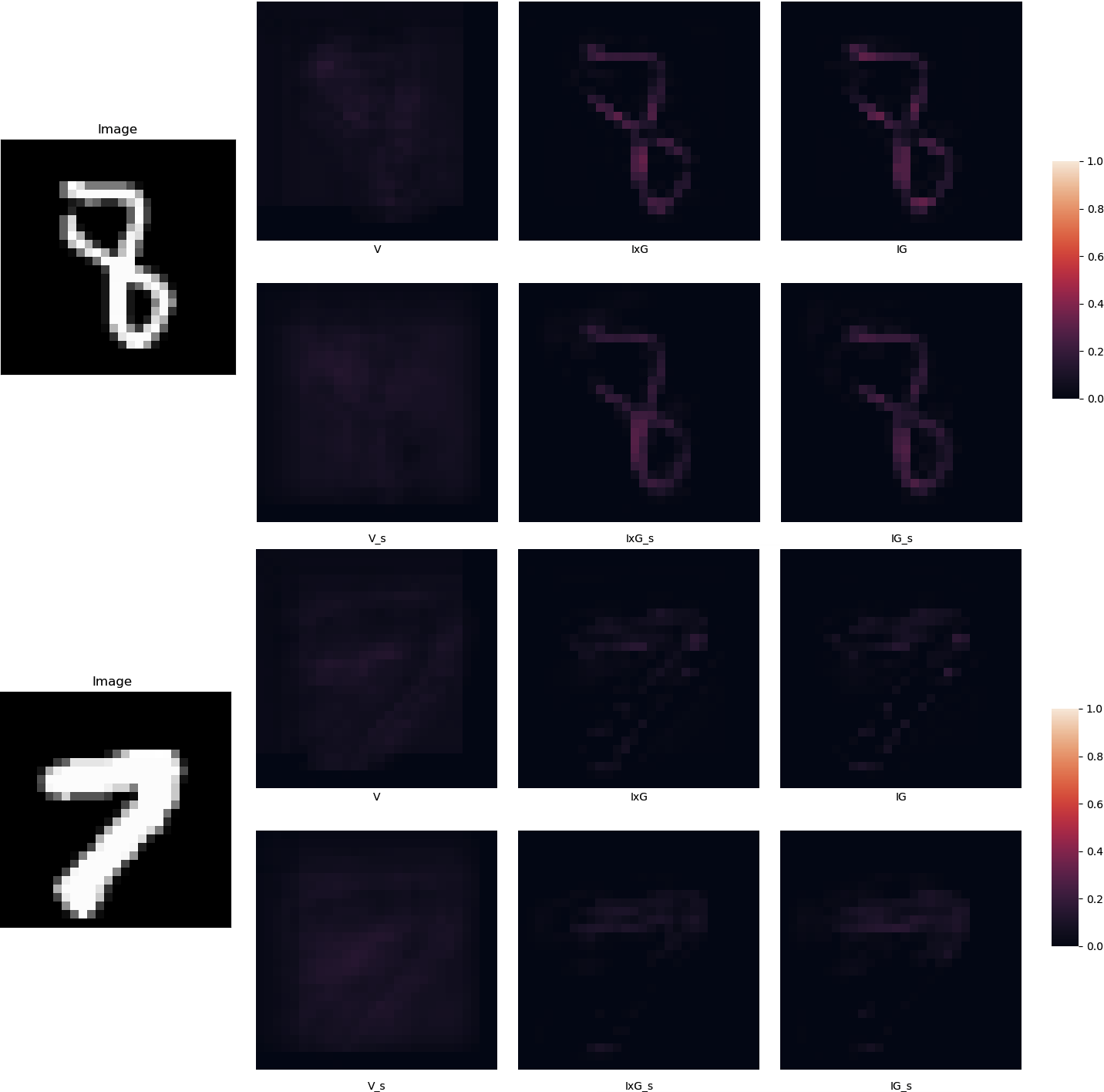}
    \caption{Visualization of the three original images and their corresponding reconstructed explanations learned from Vanilla Gradients, Input$\times$Gradients, Integrated Gradients and their SmoothGrad versions. For clearer visual effects, please increase the monitor brightness when viewing.}
    \label{fig:qualitative_smooth}
    \end{centering}
\end{figure*}

Again, the complementary results of the experiments on SmoothGrad are presented in Table \ref{tab:smooth}. We show two groups of visualization samples in Fig. \ref{fig:qualitative_smooth}, including Vanilla Gradients, Input$\times$Gradients, Integrated Gradients and their corresponding SmoothGrad versions.

\begin{table}[]
\centering
\begin{tabular}{c|cc|cc|cc}
\hline
                           & $V$     & $V\_s$            & $I\times G$      & $I\times G\_s$    & $IG$             & $IG\_s$           \\ \hline
$\overline{TA}$            & $0.541$ & $\mathbf{0.629}$ & $\mathbf{0.515}$ & $0.488$          & $\mathbf{0.543}$ & $0.508$          \\
$\overline{SC}$            & $0.727$ & $\mathbf{0.847}$ & $0.232$          & $\mathbf{0.332}$ & $0.254$          & $\mathbf{0.351}$ \\
$\overline{PC}$            & $0.596$ & $\mathbf{0.782}$ & $0.403$          & $\mathbf{0.452}$ & $0.441$          & $\mathbf{0.473}$ \\
DL                         & $0.622$ & $\mathbf{0.752}$ & $0.383$          & $\mathbf{0.424}$ & $0.413$          & $\mathbf{0.444}$ \\ \hline
$\Delta\overline{SC_a^r}$  & $0.025$ & $\mathbf{0.030}$ & $0.111$          & $\mathbf{0.299}$ & $0.107$          & $\mathbf{0.304}$ \\
$\Delta\overline{FID_a^r}$ & $0.364$ & $\mathbf{0.654}$ & $0.401$          & $\mathbf{0.681}$ & $0.378$          & $\mathbf{0.656}$ \\
VP                         & $0.389$ & $\mathbf{0.684}$ & $0.512$          & $\mathbf{0.980}$ & $0.485$          & $\mathbf{0.960}$ \\ \hline
$S_{EM}$                   & $0.241$ & $\mathbf{0.514}$ & $0.196$          & $\mathbf{0.415}$ & $0.200$          & $\mathbf{0.426}$ \\ \hline
\end{tabular}
\caption{Detailed quantitative results of the proposed evaluation method for Vanilla Gradients, Input$\times$Gradients, Integrated Gradients and their SmoothGrad versions.}
\label{tab:smooth}
\end{table}

\end{document}